\pdfoutput=1
\documentclass[11pt]{article}
\usepackage[final]{ACL2023}

\usepackage{subcaption}
\usepackage{times}
\usepackage{latexsym}
\usepackage[most,breakable]{tcolorbox}
\usepackage{stfloats}
\usepackage{booktabs}
\usepackage{float}
\usepackage{graphicx}
\tcbuselibrary{listingsutf8}

\tcbset{
    conversation/.style={
        colback=gray!10,
        colframe=gray!50,
        left=2mm,
        right=2mm,
        boxrule=0.5mm,
        arc=4mm,
        outer arc=2mm,
        top=1mm,
        bottom=1mm,
        width=\linewidth,
        fonttitle=\bfseries,
        breakable
    },
    header/.style={
        colback=gray!20,
        colframe=black,
        boxrule=0.5mm,
        arc=2mm,
        outer arc=1mm,
        width=\linewidth,
        fonttitle=\bfseries
    }
}

\usepackage[T1]{fontenc}

\usepackage[utf8]{inputenc}

\usepackage{microtype}
\usepackage{inconsolata}
\usepackage{amsfonts}
\usepackage{amsmath}

\title{CORE: Measuring Multi-Agent LLM Interaction \\ Quality under Game-Theoretic Pressures}

\author{
  Punya Syon Pandey\textsuperscript{\normalfont1,2}\quad
  Yongjin Yang\textsuperscript{\normalfont1,2}\quad
  Jiarui Liu\textsuperscript{\normalfont3} \quad
  Zhijing Jin\textsuperscript{\normalfont1,2,4}\\
  \textsuperscript{1}University of Toronto \quad
  \textsuperscript{2}Vector Institute \quad
  \textsuperscript{3}LTI, CMU \quad \\
  \textsuperscript{4}Max Planck Institute for Intelligent Systems, Tübingen, Germany \quad
  \\
  \texttt{punya.pandey@mail.utoronto.ca {} zjin@cs.toronto.edu}
}

\makeatletter
\def\blfootnote{\gdef\@thefnmark{}\@footnotetext}
\makeatother

\begin{document}
\maketitle
\begin{abstract}

Game-theoretic interactions between agents with Large Language Models (LLMs) have revealed many emergent capabilities, yet the \textit{linguistic diversity} of these interactions has not been sufficiently quantified. In this paper, we present the Conversational Robustness Evaluation Score: CORE, a metric to quantify the effectiveness of language use within multi-agent systems across different game-theoretic interactions. CORE integrates measures of cluster entropy, lexical repetition, and semantic similarity, providing a direct lens of dialog quality. We apply CORE to pairwise LLM dialogs across competitive, cooperative, and neutral settings, further grounding our analysis in Zipf's and Heaps' Laws to characterize word frequency distributions and vocabulary growth. Our findings show that cooperative settings exhibit both steeper Zipf distributions and higher Heap exponents, indicating more repetition alongside greater vocabulary expansion. In contrast, competitive interactions display lower Zipf and Heaps exponents, reflecting less repetition and more constrained vocabularies. These results provide new insights into how social incentives influence language adaptation, and highlight CORE as a robust diagnostic for measuring linguistic robustness in multi-agent LLM systems. \blfootnote{Our code is available at \url{https://github.com/psyonp/core}.}

\end{abstract}

\section{Introduction}

Human language and communication has evolved across centuries of social and evolutionary pressures. With the rise of artificial intelligence, the emergence of structured language in LLMs provides a unique opportunity to explore the underlying dynamics of linguistic evolution and communication from a novel perspective. LLM agents offer a controlled, scalable environment in which we can study how interactional pressures shape language use in real-time. Among the most compelling questions is how these agents' behaviors, driven by game-theoretic incentives \cite{hua2024gametheoreticllmagentworkflow, mao2024alympicsllmagentsmeet,Akata_2025}, drive or erode linguistic diversity \cite{NEURIPS2020_7520fa31,bouchacourt2018agentsthingsvisualrepresentations}, potentially leading to mode collapse \cite{hamilton2024detectingmodecollapselanguage}, excessive lexical repetition, and reduced semantic variation. In multi-agent systems, these incentives could range from collaboration to competition, each imposing different constraints on communication strategies and linguistic structures. 

In natural language, empirical laws such as Zipf's Law \cite{zipf1949human} and Heaps' Law \cite{10.5555/539986} have long served as foundational frameworks for understanding word frequency distributions and vocabulary growth. Zipf's Law posits an inverse relationship between word frequency and rank in a corpus, while Heaps' Law models the relationship between vocabulary size and the number of tokens produced. These laws have been observed in natural and artificial languages, offering insights into the efficiency of language use \cite{ferrericancho2001small}. However, the influence of such laws in game-theoretic interactional dynamics---particularly in multi-agent settings \cite{davidson2024evaluatinglanguagemodelagency, zhang2024chainagentslargelanguage,piatti2024cooperatecollapseemergencesustainable}---on the erosion of linguistic structure shifts has received comparatively less attention.

In this work, we introduce CORE: a unified metric for multi-agent LLM Conversational Robustness Evaluation. CORE integrates cluster entropy, lexical repetition, and semantic stagnation to provide a direct measure of linguistic diversity across dialogs. We simulate dialogs between pairs of LLM agents under competitive, cooperative, and neutral game-theoretic conditions, applying Zipf’s and Heaps’ analyses to interpret the patterns CORE detects. Our findings reveal that social incentives shape lexical diversity and repetition: cooperative settings tend to expand vocabulary while also increasing repetition, whereas competitive settings compress both vocabulary diversity and conversational modes. To our knowledge, this is the first large-scale study to jointly analyze statistical linguistic laws and conversational diversity metrics in game-theoretic multi-agent LLM interactions.

\section{Related Work}

\begin{figure*}[t]
    \centering
    \includegraphics[width=1\linewidth]{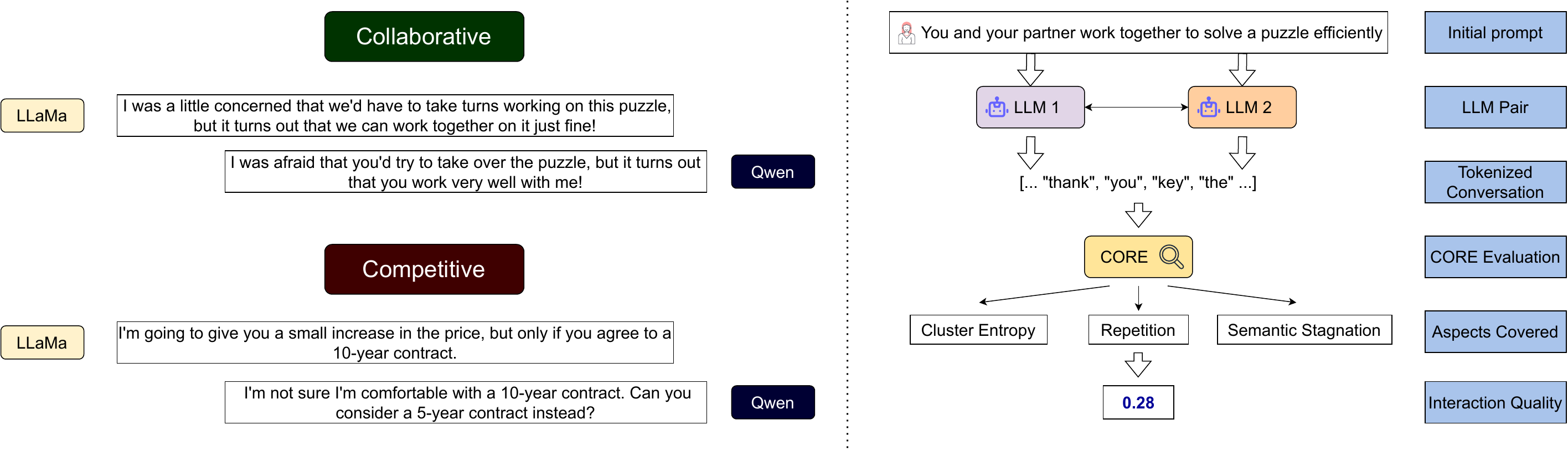}
    \caption{Illustration of CORE’s application domain and workflow for efficiently assessing linguistic diversity.}
    \label{fig:wide_placeholder}
    
\end{figure*}

\paragraph{Game Theory and Language Evolution}
Game-theoretic frameworks have long been used to model the emergence and evolution of communication systems, both in human and artificial settings. Foundational work in evolutionary linguistics explores how signaling systems emerge under coordination pressures \cite{SMITH2010231, Hayes2014Cooperation, doi:10.1126/science.291.5501.114}. In artificial environments, multi-agent reinforcement learning (MARL) has shown that structured communication protocols can emerge when agents interact to maximize shared or individual rewards \cite{lazaridou2017multiagentcooperationemergencenatural, jaques2019socialinfluenceintrinsicmotivation}. Recent work has extended these paradigms to LLMs, highlighting their capacity to exhibit strategic and socially grounded behaviors under cooperative and adversarial setups \cite{hua2024gametheoreticllmagentworkflow, mao2024alympicsllmagentsmeet, Akata_2025}. However, these studies primarily emphasize behavioral alignment or task success, often overlooking the underlying linguistic structure of the generated communication. We address this gap by proposing a unified framework that measures conversational diversity, lexical repetition, and semantic richness—capturing how strategic incentives shape the quality of emergent language

\paragraph{Statistical Laws of Language}
Zipf’s Law \cite{zipf1949human} and Heaps’ Law \cite{10.5555/539986} provide robust empirical tools for analyzing frequency-rank distributions and vocabulary growth, respectively. These regularities are interpreted as reflections of communicative efficiency and cognitive constraints \cite{ferrericancho2001small, Piantadosi2014Zipf}. In artificial agents, studies have shown that symbolic communication protocols can display statistically-defined behavior under certain optimization conditions \cite{chaabouni-etal-2020-compositionality, bouchacourt2018agentsthingsvisualrepresentations}. However, these investigations are often restricted to synthetic languages, limited vocabularies, or visual environments. In contrast, we apply these statistical tools to unconstrained natural language generated by open-source LLMs in interactive settings. In our framework, deviations in Zipf and Heaps parameters are combined with semantic and clustering analyses to detect lexical redundancy, vocabulary stagnation, and mode collapse \cite{hamilton2024detectingmodecollapselanguage} in multi-agent communication.

\paragraph{LLMs in Multi-Agent Environments}
Recent efforts have explored LLMs in interactive multi-agent setups, including debate \cite{liang-etal-2024-encouraging, zhang2024llmsbeathumansdebating}, collaborative decision-making \cite{tran2025multiagentcollaborationmechanismssurvey, shen2024learningdecodecollaborativelymultiple, zhu2025multiagentbenchevaluatingcollaborationcompetition}, and social simulation \cite{Argyle_2023, tang2025gensimgeneralsocialsimulation}. These works often focus on alignment, role consistency, or behavioral coherence, with relatively little attention paid to the statistical properties of the language produced during interaction. Moreover, some studies evaluate interactions systematically across a taxonomy of incentives (e.g., cooperation vs. competition) or assess structural linguistic outcomes at scale \cite{piatti2024cooperatecollapseemergencesustainable,zhao2024competeaiunderstandingcompetitiondynamics}. Our study is the first to evaluate how cooperative, competitive, and neutral settings directly modulate the linguistic statistics of interactions between multiple open-source LLMs. This approach bridges a key gap, revealing how strategic incentives shape not just agent behavior but also fundamental patterns in language.

\section{Preliminaries}

\subsection{Problem Definition}

Consider a multi-agent system \(\mathcal{M} = \{m_1, m_2\}\) interacting through dialogs represented as sequences of utterances. We define a dataset of dialogs \(\mathcal{D} = \{d_1, d_2, \ldots, d_N\}\), where each dialog 
\[
d_i = (u_1^i, u_2^i, \ldots, u_{T_i}^i)
\]
consists of utterances generated by agents according to a conditional probability distribution:
\[
u_t^i \sim P(u_t \mid u_{<t}^i, m)
\]
where \(u_{<t}^i = (u_1^i, \ldots, u_{t-1}^i)\) is the dialog history and \(m \in \mathcal{M}\) is the generating agent. The interaction dynamics induce a distribution over dialog trajectories and corresponding latent conversational modes \(\mathcal{C} = \{c_1, c_2, \ldots, c_K\}\), representing distinct clusters of behavior or thematic patterns within the dialogs. A key challenge arises when the empirical distribution over modes \(P_C\) becomes concentrated on a small subset of \(\mathcal{C}\), leading to \emph{mode collapse}. This concentration indicates a loss of diversity in multi-agent interactions, often accompanied by increased lexical repetition and semantic similarity between utterances, which degrade dialog quality.

\paragraph{Objective} We aim to develop a comprehensive, theoretically grounded metric that characterizes and quantifies these phenomena—mode collapse, repetition, and semantic stagnation—in multi-agent dialog data. The framework will enable systematic evaluation and comparison of dialog quality without dependence on external task rewards or annotations within LLM interactions.

\subsection{Statistical Laws of Language}

\paragraph{Zipf's Law}
Zipf's Law \cite{zipf1949human} is an empirical law stating that the frequency $f(w)$ of a word $w$ is inversely proportional to its rank $r(w)$ when words are sorted by descending frequency:
\begin{equation}
f(w) \propto \frac{1}{r(w)^\alpha}, \quad \alpha \approx 1
\end{equation}

This results in a power-law distribution over word frequencies. In natural language corpora, this skewed distribution implies that a small subset of tokens dominates usage, which has implications for model capacity in multi-agent interactions.

\begin{table*}[b]
    \centering
    \small
    \resizebox{0.8\linewidth}{!}{
    \begin{tabular}{ll}
        \toprule
        \textbf{Mode} & \textbf{Seed Prompt} \\
        \midrule
        Cooperative & You and your partner work together to solve a puzzle efficiently \\
        Competitive & You are competing in a negotiation and want to outwit and outperform your opponent \\
        Neutral & You engage in casual, open-ended conversation with no specific agenda \\
        \bottomrule
    \end{tabular}
    }
    \caption{Initial prompts used to elicit model behavior across different game-theoretic interaction modes.}
    \label{tab:sample}
\end{table*}

\paragraph{Heap's Law}
Heap's Law \cite{10.5555/539986} describes the growth of the number of unique word types $V(n)$ as a function of the total number of word tokens $n$:
\begin{equation}
V(n) = K n^\beta, \quad 0 < \beta < 1
\end{equation}
where $K$ and $\beta$ are empirical constants determined by the corpus. This law captures the sublinear increase of vocabulary size as data scales, which is central to understanding lexical diversity, generalization behavior, and the challenges of open-vocabulary modeling.

\subsection{Game-Theoretic Conditions}
We define a game $\mathcal{G} = (N, \{S_i\}, \{u_i\})$ consisting of $N$ agents, where each agent $i \in \{1, \ldots, N\}$ selects a strategy $s_i \in S_i$ to maximize a utility function $u_i: \prod_j S_j \rightarrow \mathbb{R}$. We consider three canonical interaction modes:
\begin{itemize}
    \item \textbf{Cooperative:} $u_i = u_j$ for all $i, j$, with agents jointly optimizing a shared utility function.
    \item \textbf{Competitive:} $u_i \neq u_j$, and agents have adversarial objectives, often maximizing utility at the other's expense.
    \item \textbf{Neutral:} Agents act independently with unaligned or orthogonal utility functions, without explicit cooperation or conflict.
\end{itemize}

These modes characterize the structural conditions under which agents interact, make decisions, or exchange information. In multi-agent systems, these distinctions help formalize learning dynamics, reward alignment, and coordination strategies.

\section{Methodology}

In this section, we introduce the CORE metric alongside the experimental setup and our model selection. Together, these elements define our evaluation framework and ensure comparability across different multi-agent configurations.

\subsection{CORE}

To quantitatively evaluate the diversity of multi-agent dialogs, we propose the \textbf{Conversational Robustness Evaluation} (\textbf{CORE}) score. CORE jointly captures three aspects of internal robustness: mode collapse \cite{hamilton2024detectingmodecollapselanguage}, lexical repetition, and semantic stagnation. We denote by \(K_{\max}\) the maximum modes for normalization. We then define the \(\textbf{CORE} \in [0,1]\) metric as following:

\noindent
where

\begin{equation}
H(C) = -\sum_{i=1}^K p_i \log p_i
\tag{4}
\end{equation}

is the Shannon entropy of the mode distribution. Let \(\mathcal{Z}\) be the multiset of all \(n\)-grams extracted from the dialogs, defined as

\begin{equation}
\mathcal{Z} = \bigcup_{d \in \mathcal{D}} \{ z \in \mathrm{ngrams}(d, n) \}
\tag{5}
\end{equation}

where \(\mathrm{count}(z)\) is the frequency of the \(n\)-gram \(z\) in \(\mathcal{Z}\). For each utterance \(u_j\), let \(\mathbf{e}_j \in \mathbb{R}^d\) denote its embedding vector (e.g., derived via Sentence-BERT). The cosine similarity between consecutive utterance embeddings is given by

\begin{equation}
\cos(\mathbf{e}_j, \mathbf{e}_{j+1}) = \frac{\mathbf{e}_j \cdot \mathbf{e}_{j+1}}{\|\mathbf{e}_j\| \, \|\mathbf{e}_{j+1}\|}
\tag{6}
\end{equation}

We introduce hyperparameters $\alpha, \beta > 0$, corresponding to the empirical Zipf and Heaps exponents of the target corpus derived from generated dialog data, respectively. This design choice allows the CORE metric to account for inherent linguistic characteristics of each LLM, ensuring a nuanced and generalizable evaluation of a model's internal robustness by assessing how much its generated output deviates from its own typical statistical properties under various interaction conditions. 

\begin{figure*}[t]
\resizebox{\textwidth}{!}{
\fbox{
$\displaystyle
\mathrm{CORE}(\mathcal{D}) = \underbrace{\frac{H(C)}{\log K_{\max}}}_{\text{Normalized Mode Entropy}} \times
\underbrace{\left(1 - \frac{\sum_{z \in \mathcal{Z}} \mathbf{1}(\mathrm{count}(z) > 1)}{|\mathcal{Z}|}\right)^{\alpha}}_{\text{Repetition Penalty}} \times
\underbrace{\left(1 - \frac{1}{N-1} \sum_{j=1}^{N-1} \cos(\mathbf{e}_j, \mathbf{e}_{j+1})\right)^{\beta}}_{\text{Semantic Stagnation Penalty}}
\hfill (3)$
}
}
\label{fig:core_formula}
\end{figure*}

\subsection{Model Selection}
We employ eight open-source LLMs spanning several architectures for a thorough assessment of game-theoretic incentives in shaping language structure within current LLMs. Specifically, we consider Llama-3.1 8B \cite{meta2025llama3}, Llama-3.1-8B Instruct \cite{meta2025llama3.1}, Gemma-7B \cite{gemmateam2024gemmaopenmodelsbased}, Gemma-7B Instruct \cite{gemmateam2024gemmaopenmodelsbased}, Qwen-3-8B \cite{yang2025qwen3technicalreport}, Qwen-2.5-7B Instruct \cite{qwen2025qwen25technicalreport}, Mistral-7B v03 \cite{jiang2023mistral7b}, and Mistral-7B Instruct \cite{jiang2023mistral7b}. 

\begin{figure*}[t]
    \centering    
    \includegraphics[width=\linewidth]{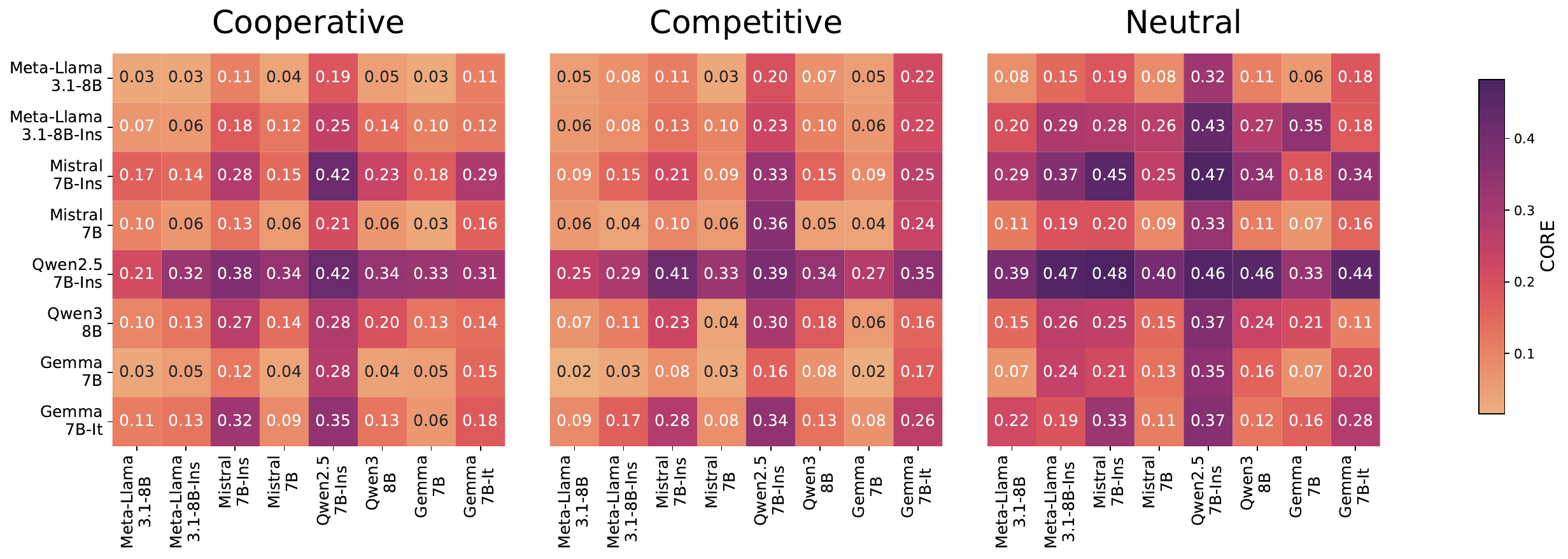}
    \caption{CORE values across model-pair interactions under cooperative, competitive, and neutral conditions. Higher values indicate more lexically diverse interactional quality and varied vocabulary usage within dialogs.}
    \label{fig:yippyipp}
\end{figure*}

\subsection{Experimental Setup}
Each $(\text{agent pair}, \text{condition})$, is evaluated on 30 dialogs of 10 alternating turns, starting from a condition-specific prompt (Table~\ref{tab:sample}). Generation uses nucleus sampling (temperature 0.7, top-p 0.9) with a 128-token limit. All utterances are concatenated and tokenized using a case-insensitive regex. We compute the Zipf $\alpha$ and Heap's $\beta$ to analyze frequency concentration and vocabulary growth, and apply the Mann-Whitney U test \cite{doi:https://doi.org/10.1002/9780470479216.corpsy0524} for statistical significance. The evaluation covers 64 pairs × 3 conditions × 30 dialogs = 5,760 interactions. Full details are in Appendix~\ref{appendix:experiments}.

\section{Experiments and Results}

This section centers on evaluating multi-agent interactions using our proposed CORE metric. We also contextualize CORE’s findings with analyses of token and rank-frequency distributions, Zipf’s and Heaps’ laws, and qualitative dialog dynamics, demonstrating CORE’s efficacy in capturing interaction quality.

\subsection{Quantifying Interaction Mode Collapse}

As part of our analysis, we calculate the CORE results between model pairs and game-theoretic interaction modes (Figure~\ref{fig:yippyipp}). Our results consistently show that neutral dialogs exhibit the highest CORE values, while cooperative and competitive dialogs show progressively lower values. This reflects the degree of mode collapse \cite{hamilton2024detectingmodecollapselanguage} across each interaction mode, examining cluster entropy, repetition, and semantic similarity. Additional summary statistics are in Table~\ref{summary}.

\subsection{Token and Rank-Frequency Distribution}

\paragraph{Token Analysis}
To answer gain insight into lexical variation across interaction settings, we examine the distribution of unique tokens generated under cooperative, competitive, and neutral conditions (Table~\ref{summary}). Cooperative dialogs exhibit the lowest lexical diversity, reusing a narrower vocabulary—consistent with goal-oriented repetition. Furthermore, competitive interactions show a moderately broader range of unique tokens, suggesting underlying dynamics that incentivize variation. Neutral settings display the highest lexical diversity, suggesting more open-ended conversational goals and a reduced need for strategic lexical alignment.

\begin{figure}[H]
    \centering

    \begin{subfigure}[b]{0.48\linewidth}
        \includegraphics[width=\linewidth]{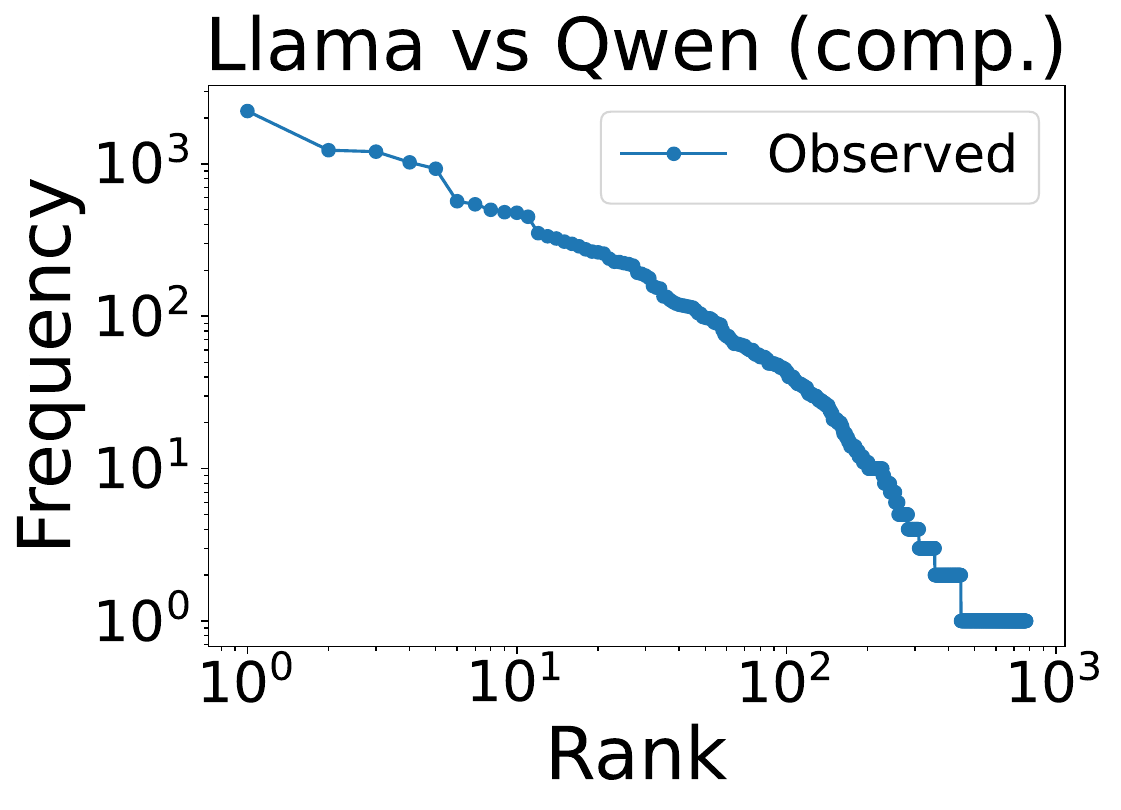}
        \label{fig:1a}
    \end{subfigure}
    \hfill
    \begin{subfigure}[b]{0.49\linewidth}
        \includegraphics[width=\linewidth]{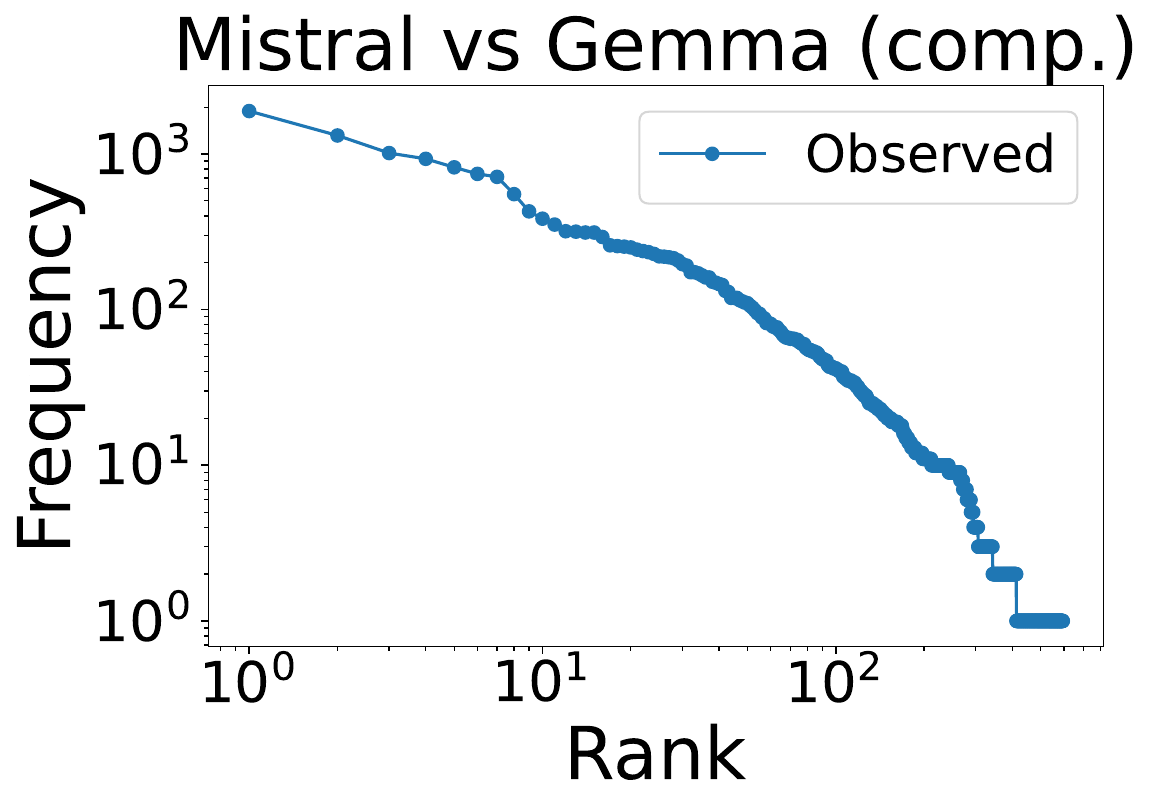}
        \label{fig:1b}
    \end{subfigure}

    \vspace{0.5em}

    \begin{subfigure}[b]{0.48\linewidth}
        \includegraphics[width=\linewidth]{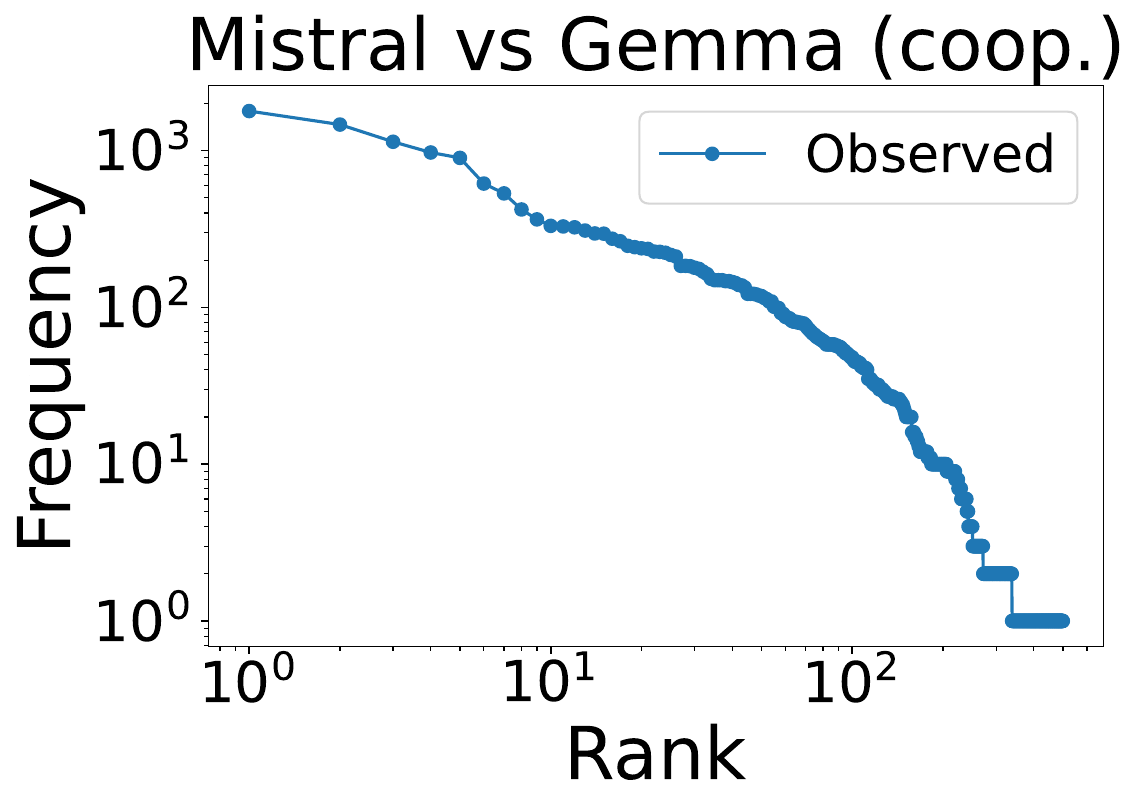}
        \label{fig:1c}
    \end{subfigure}
    \hfill
    \begin{subfigure}[b]{0.49\linewidth}
        \includegraphics[width=\linewidth]{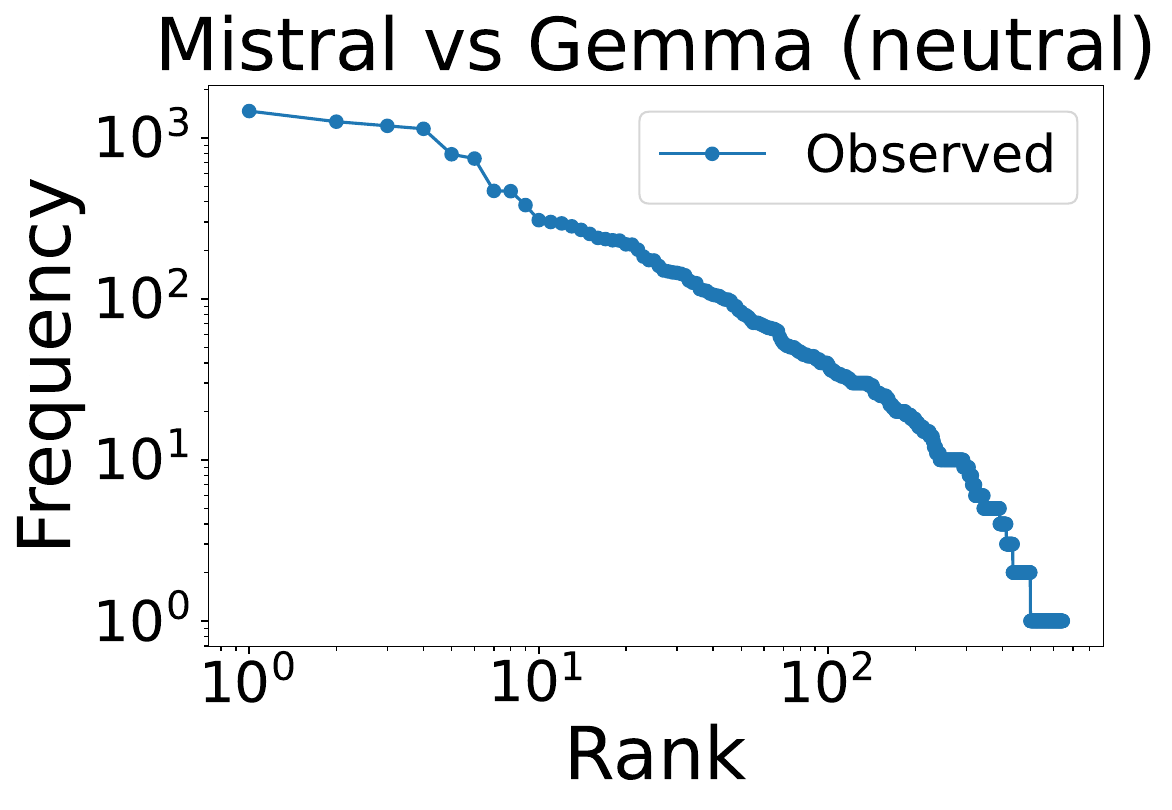}
        \label{fig:1d}
    \end{subfigure}

\caption{Zipfian behavior across models and modes signals linguistic efficiency in multi-agent settings.}
 \label{fig:grid}
\end{figure}

\paragraph{Rank-Frequency Distribution}
As an extension of token analysis, we examine rank-frequency distributions aggregated across all dialog outputs for each setting. Figure~\ref{fig:grid} shows examples confirming that generated language across modes follows Zipfian structure to varying degrees, but the slope and curvature differ substantially by condition. These effects are most pronounced in agent pairs where both models are instruction-tuned, suggesting alignment objectives may interact non-trivially with incentive structures to impact lexical structure.

\subsection{Multi-Agent Lexical Distributions}

\begin{figure*}[t]
    \centering
    \includegraphics[width=\linewidth]{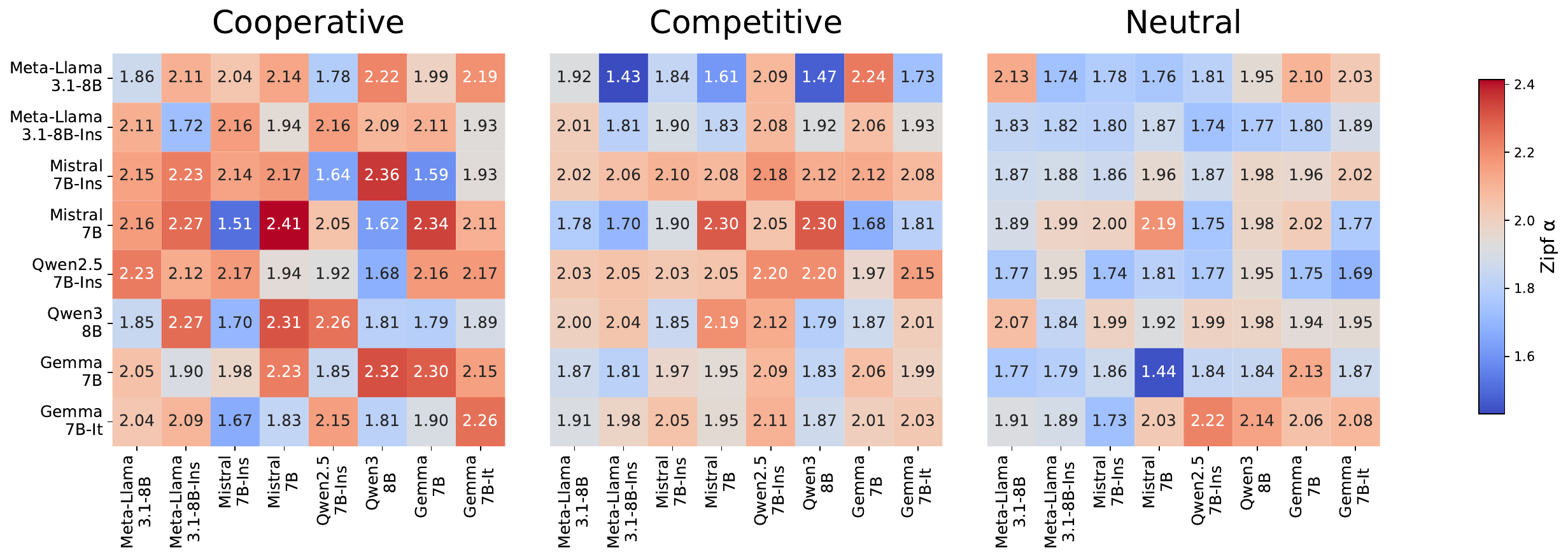}
    \caption{Zipf $\alpha$ exponents across model-pair interactions. Higher $\alpha$ indicates stronger frequency concentration among high-rank tokens, while lower $\alpha$ reflects flatter distributions with higher lexical dispersion.}
    \label{fig:zipf1}
    
    \includegraphics[width=\linewidth]{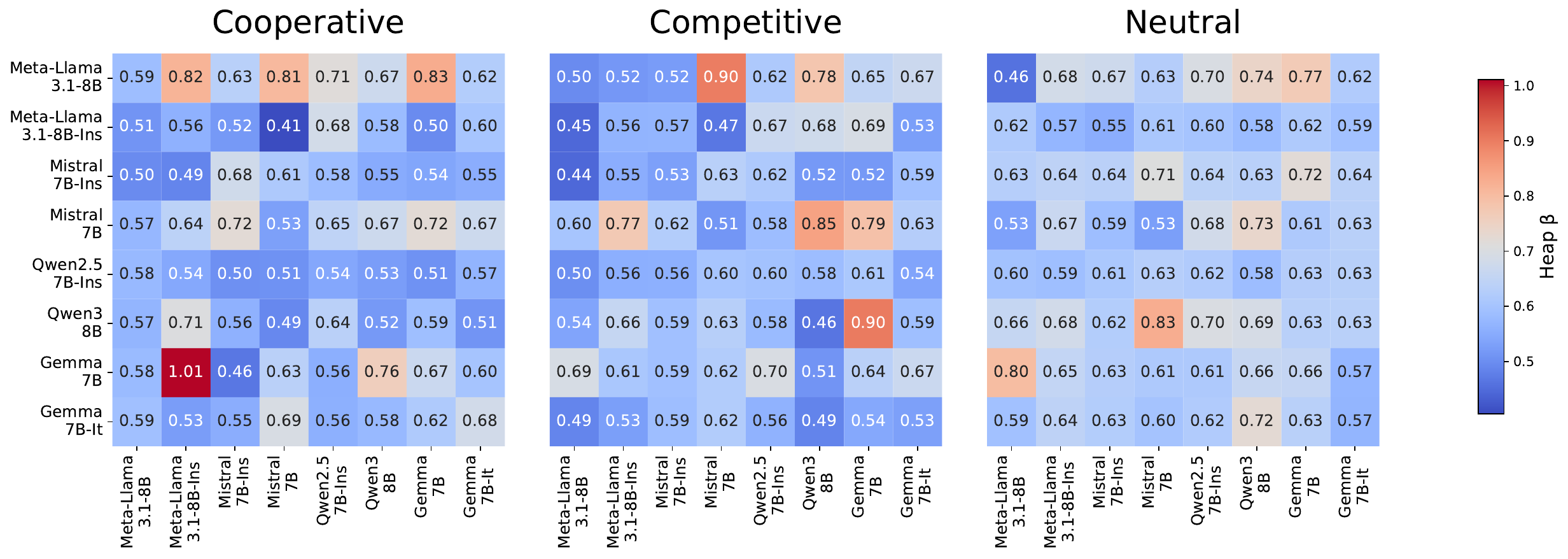}
    \caption{Heap $\beta$ exponents across model-pair interactions. The exponent $\beta$ reflects the rate of vocabulary growth as a function of dialog length, with higher values indicating greater lexical diversity}
    \label{fig:heap1}

    \vspace{0.5cm}
\end{figure*}

We examine Zipf and Heaps values across model pairs and interaction modes within the CORE framework (Figures~\ref{fig:zipf1} and~\ref{fig:heap1}). Cooperative dialogs exhibit higher $\alpha$ and $\beta$, indicating frequent reuse of dominant tokens and vocabulary expansion, while competitive and neutral dialogs show lower $\alpha$ and $\beta$, aligning with CORE’s diversity–repetition trade-offs. Additional metrics are listed in Table~\ref{summary}.

\begin{table}[H]
\centering
\resizebox{0.45\textwidth}{!}{
\begin{tabular}{lccccc}
\toprule
\textbf{Condition} & \textbf{Mean} & \textbf{Std Dev} & \textbf{Max} & \textbf{Min} & \textbf{Range} \\
\midrule
\multicolumn{6}{c}{\textbf{CORE}} \\
\addlinespace
Cooperative & 0.1630 & 0.1578 & 0.7752 & 0.0 & 0.7752 \\
Competitive & 0.1548 & 0.1568 & 0.6631 & 0.0 & 0.6631 \\
Neutral     & 0.2470 & 0.1884 & 0.8656 & 0.0 & 0.8656 \\
\addlinespace
\midrule
\multicolumn{6}{c}{\textbf{Zipf Exponent}} \\
\addlinespace
Cooperative & 2.0323 & 0.2131 & 2.4142 & 1.5139 & 0.9003 \\
Competitive & 1.9716 & 0.1728 & 2.3004 & 1.4317 & 0.8687 \\
Neutral     & 1.8985 & 0.1370 & 2.2202 & 1.4439 & 0.7763 \\
\addlinespace
\midrule
\multicolumn{6}{c}{\textbf{Heap Exponent}} \\
\addlinespace
Cooperative & 0.6036 & 0.1008 & 1.0111 & 0.4053 & 0.6058 \\
Competitive & 0.5995 & 0.0979 & 0.9013 & 0.4440 & 0.4574 \\
Neutral     & 0.6368 & 0.0614 & 0.8286 & 0.4590 & 0.3697 \\
\midrule
\multicolumn{6}{c}{\textbf{Unique Tokens}} \\
\addlinespace
Cooperative & 1058.63 & 2110 & 372 & 1738 & 443.48 \\
Competitive & 1162.55 & 2399 & 436 & 1963 & 497.04 \\
Neutral     & 1699.34 & 3363 & 565 & 2798 & 665.31 \\
\addlinespace
\bottomrule
\end{tabular}
}

\caption{Summary statistics across CORE, Zipf’s, Heap’s, and unique token results.}
\label{summary}
\end{table}

\begin{table*}[t]
\centering
\small
\resizebox{\textwidth}{!}{
\begin{tabular}{lcccc ccccc ccccc}
\toprule
& \multicolumn{4}{c}{\textbf{Competitive}} & & \multicolumn{4}{c}{\textbf{Cooperative}} & & \multicolumn{4}{c}{\textbf{Neutral}} \\
\cmidrule{2-5} \cmidrule{7-10} \cmidrule{12-15}
\textbf{Model} & $\alpha$ & $\beta$ & Unique & CORE & & $\alpha$ & $\beta$ & Unique & CORE & & $\alpha$ & $\beta$ & Unique & CORE \\
\midrule
Llama 3.1-8B & 1.92 & 0.50 & 457 & 0.05& & 1.86 & 0.59 & 434 & 0.03& & 2.13 & 0.46 & 621 & 0.08\\
Llama 3.1-8B Instruct & 1.81 & 0.56 & \textbf{1120} & \textbf{0.08}& & 1.72 & 0.56 & \textbf{824} & \textbf{0.06}& & 1.82 & 0.57 & \textbf{1925} & 0.29\\
Mistral-7B Instruct v0.3 & 2.10 & 0.53 & \textbf{1266} & \textbf{0.21}& & 2.14 & 0.68 & \textbf{1210} & \textbf{0.28}& & 1.86 & 0.64 & \textbf{2299} & 0.45\\
Mistral-7B v0.3 & 2.30 & 0.51 & 578 & 0.06& & 2.41 & 0.53 & 436 & 0.06& & 2.19 & 0.53 & 565 & 0.09\\
Qwen 2.5-7B Instruct & 2.20 & 0.60 & \textbf{2399} & \textbf{0.39}& & 1.92 & 0.54 & \textbf{2110} & \textbf{0.42}& & 1.77 & 0.62 & \textbf{2651} & 0.46\\
Qwen 3-8B & 1.79 & 0.46 & 1467 & 0.18& & 1.81 & 0.52 & 1358 & 0.20& & 1.98 & 0.69 & 2714 & 0.24\\
Gemma-7B & 2.06 & 0.64 & 505 & 0.02& & 2.30 & 0.67 & 460 & 0.05& & 2.13 & 0.66 & 628 & 0.07\\
Gemma-7B Instruct & 2.03 & 0.53 & \textbf{922} & \textbf{0.26}& & 2.26 & 0.68 & \textbf{809} & \textbf{0.18}& & 2.08 & 0.57 & \textbf{908} & 0.28\\
\bottomrule
\end{tabular}
}
\caption{Self-play metrics across all models for competitive, cooperative, and neutral interaction modes. Instruction-tuned models generally achieve higher unique token counts and higher CORE values, indicating lexically diverse interactions with notable gains in neutral settings.}
\label{tab:heaps_8models}
\end{table*}

\subsection{Self-Play Interactions}

\begin{figure}[H]
    \centering
    \includegraphics[width=1\linewidth]{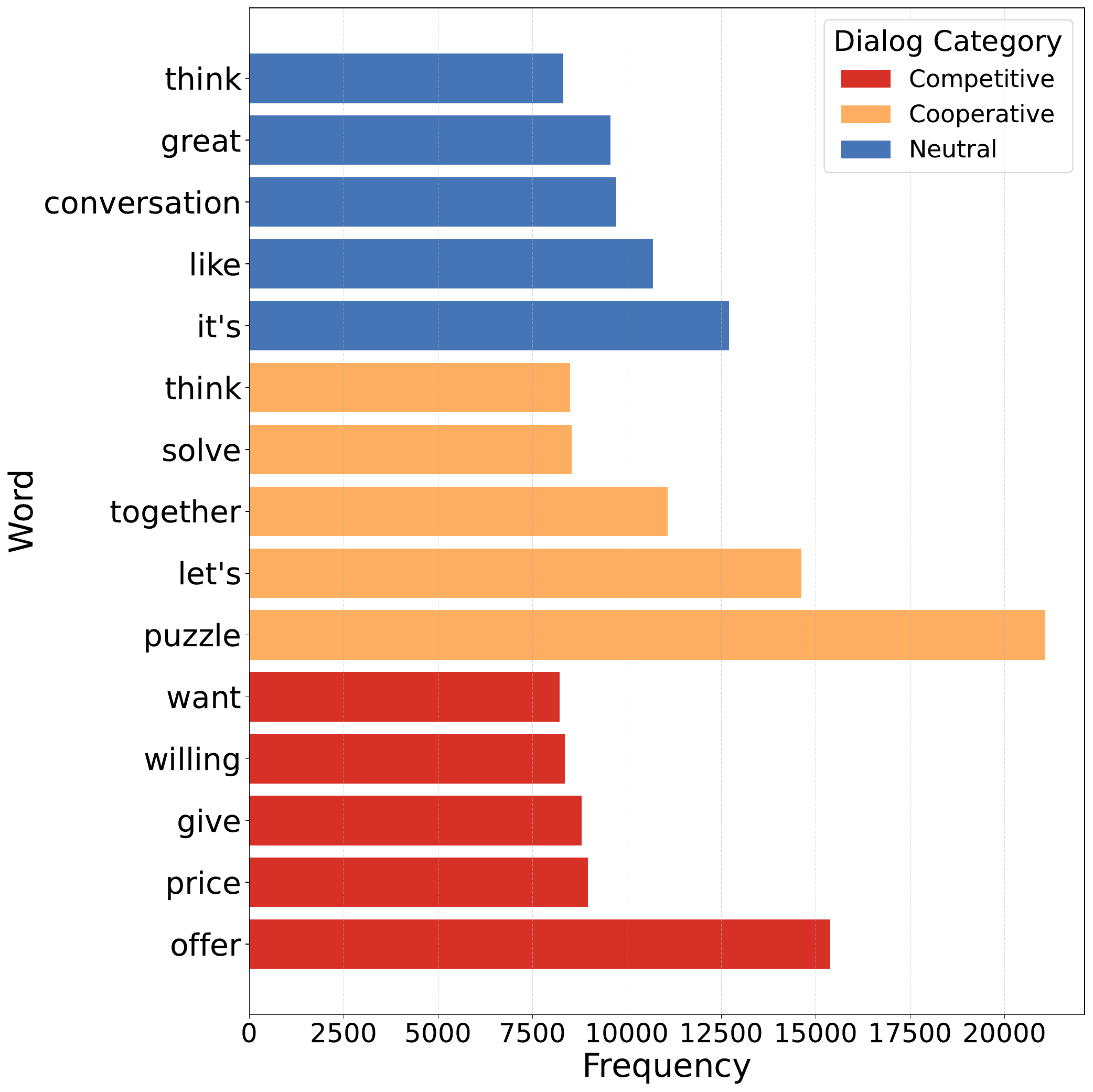}
    \caption{Top five most frequent words across competitive, cooperative, and neutral settings. The cooperative condition exhibits notably higher repetition, highlighting convergence on shared vocabulary.}
    \label{fig:haha}
\end{figure}

We compare instruction-tuned and base models across self-play settings, measuring CORE values, Zipf ($\alpha$) and Heaps ($\beta$) exponents, and unique token counts (Table~\ref{tab:heaps_8models}). Self-play yields more symmetric, internally consistent communication with generally lower $\alpha$ and $\beta$, reflecting reduced lexical diversity and vocabulary generation compared to multi-agent interactions. Cooperative settings further suppress unique token generation, as agents converge on shared vocabulary (Figure~\ref{fig:haha}). Instruction-tuned models, while showing lower $\beta$ and less lexical variety than base models in multi-agent contexts, produce more unique tokens and achieve higher CORE values in self-play, indicating that alignment objectives, though constraining vocabulary in interactive settings, can promote richer language use when the model interacts with itself. This reinforces that instruction-tuned models preserve semantic differences in self-play interactions.

\subsection{Behavioral Metrics}

\begin{figure}[H]
    \centering
    \includegraphics[width=0.72\linewidth]{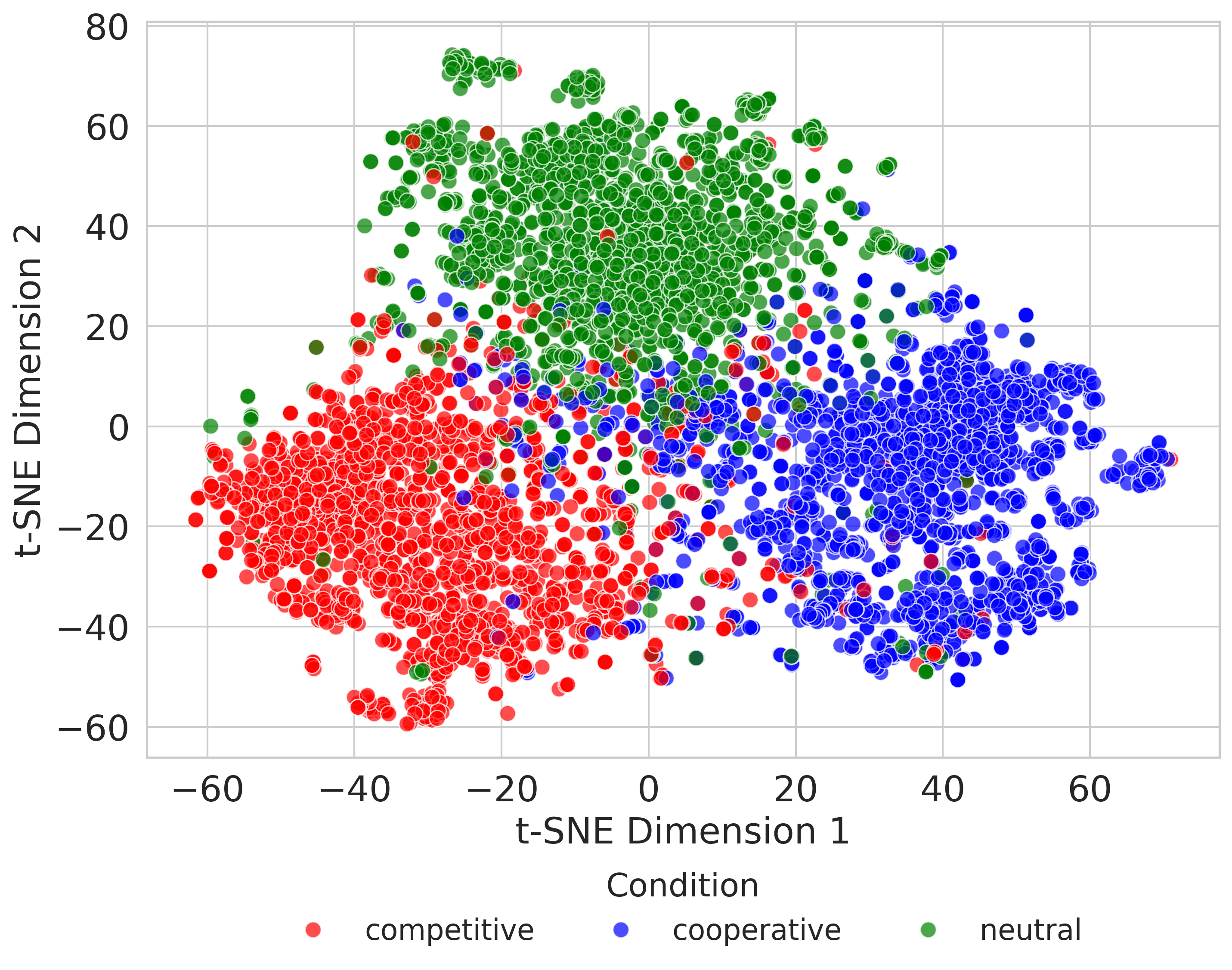}
    \caption{Latent feature projection of dialogs across game-theoretic mode, embedded using TF-IDF features and reduced via t-SNE. Clusters indicate linguistic similarity, suggesting distinct conversational styles.}
    \label{fig:tsne}
\end{figure}

We examine latent semantic patterns through t-SNE \cite{JMLR:v9:vandermaaten08a} embeddings (Figure~\ref{fig:tsne}), noticing distinct clusters across cooperative, competitive, and neutral modes. 

\begin{figure*}[b]
    \centering
    \includegraphics[width=0.89\linewidth]{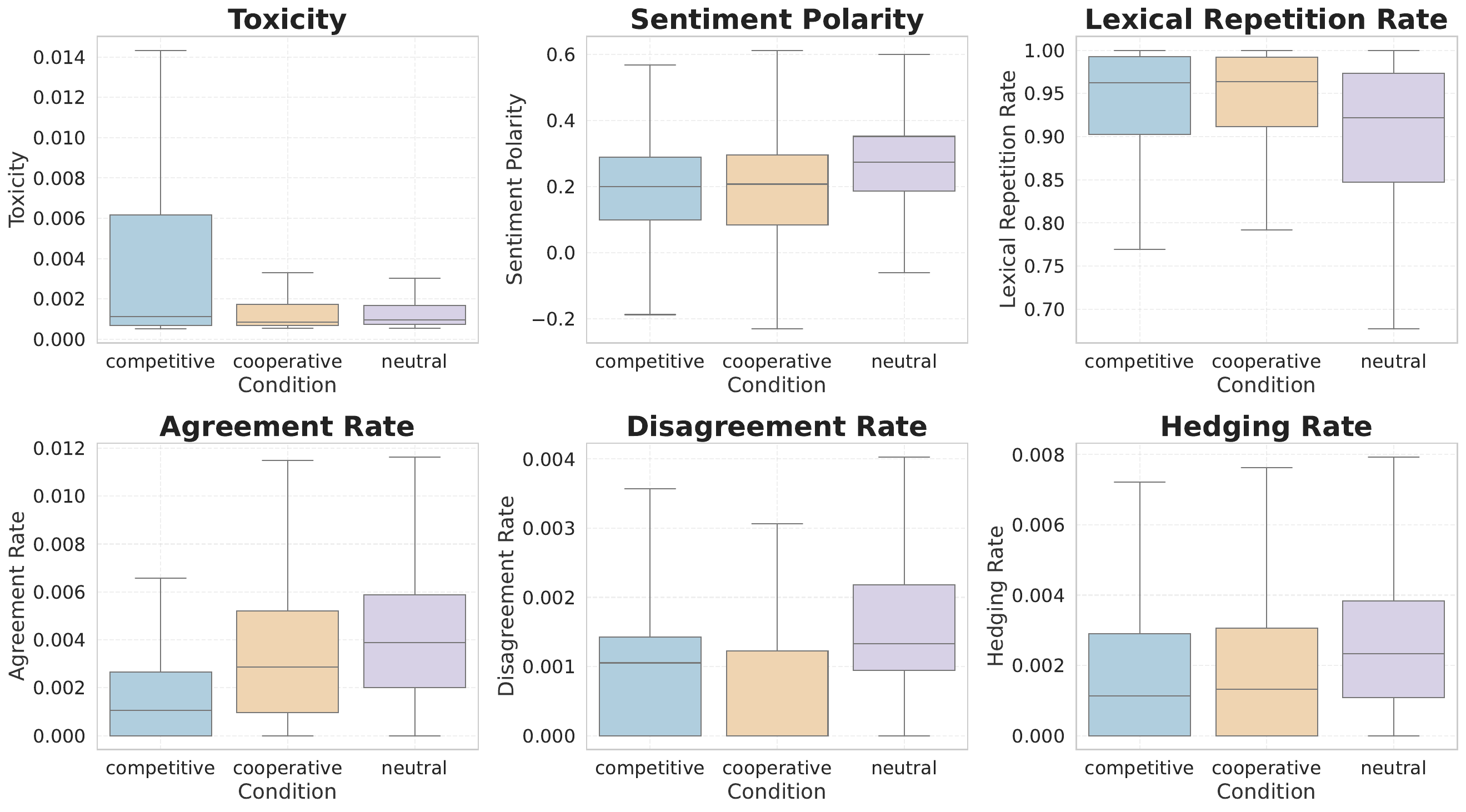}
    \caption{Behavioral metrics across all modes, showing higher repetition and toxicity across competitive dialogs.}
    \label{fig:add}
\end{figure*}

\begin{figure*}[t]
    \centering
    \includegraphics[width=\linewidth]{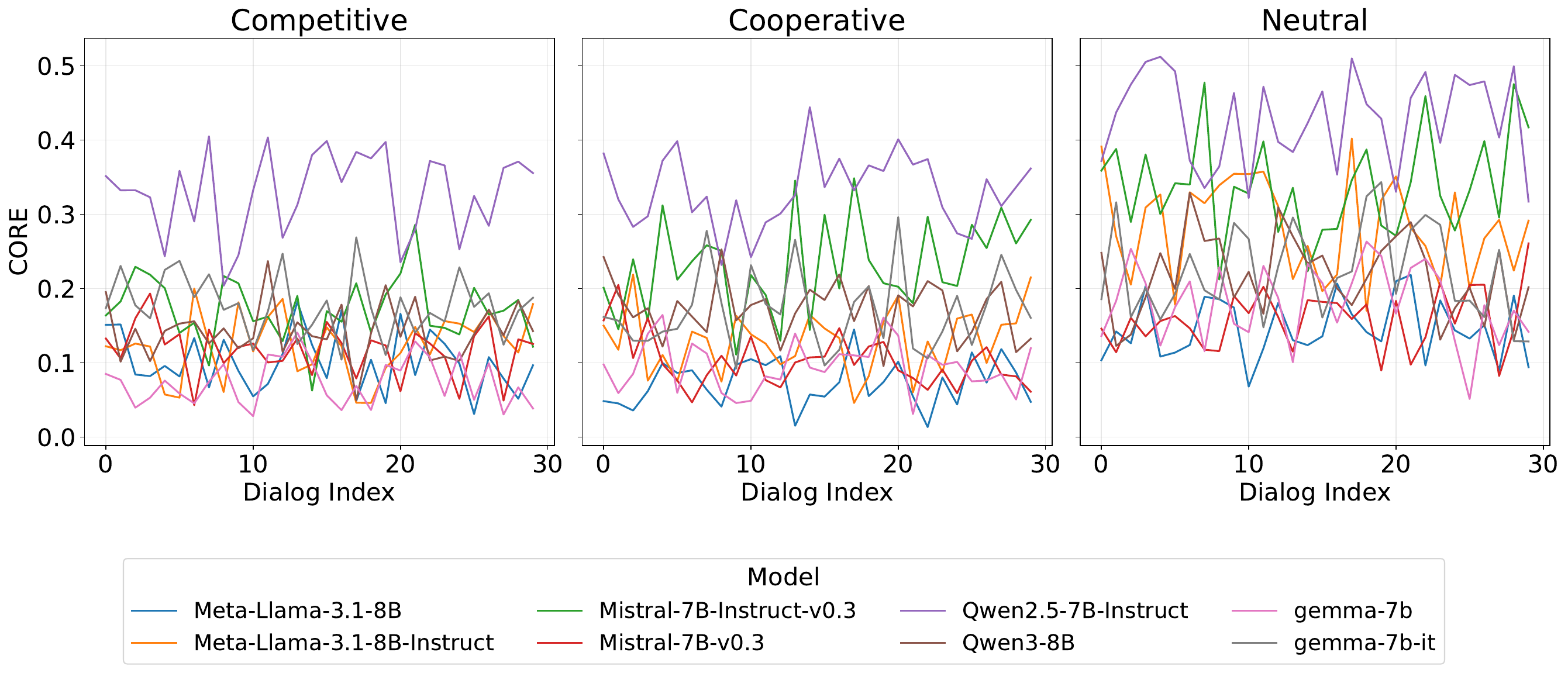}
    \caption{Temporal trends in CORE scores across dialog indexes across cooperative, competitive, and neutral conditions. Neutral dialogs consistently achieve the highest scores, with cooperative and competitive settings showing lower and more variable patterns over time.}
    \label{fig:placeholder}
\end{figure*}

\begin{figure*}[b]
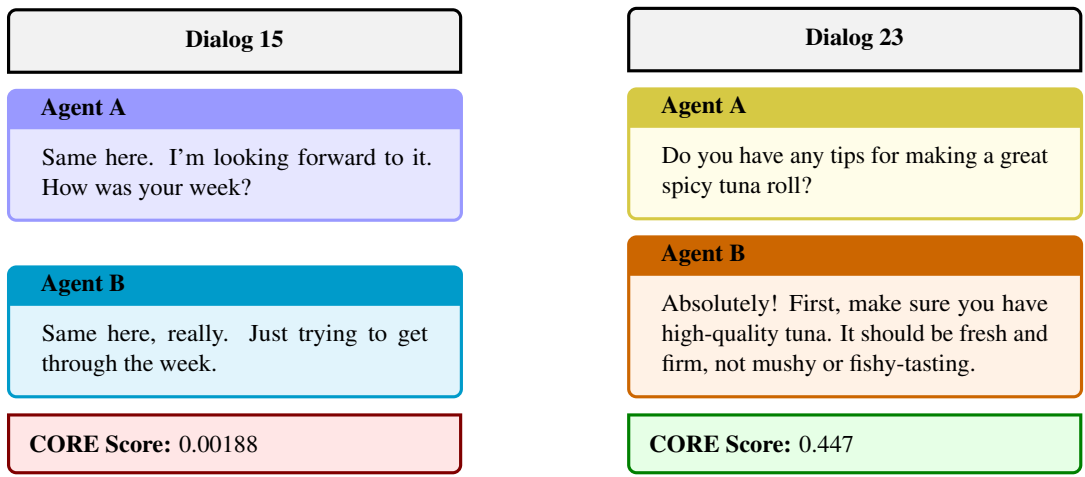

\centering
\resizebox{0.9\textwidth}{!}{ 
  \newtcolorbox[auto counter, number within=section]{dialogbox}[2][]{colframe=blue!60!black, colback=blue!5, coltitle=black, fonttitle=\bfseries, title=#2,#1}

  \begin{minipage}[b]{0.45\textwidth}
        \begin{tcolorbox}[colback=gray!10, colframe=black, sharp corners=south, boxrule=0.5mm, width=2.83in, left=2mm, right=2mm, top=2mm, bottom=2mm]
      \centering
      \textbf{Dialog 15}
      \end{tcolorbox}
      \begin{dialogbox}[colframe=blue!40!white, colback=blue!10, fonttitle=\bfseries]{Agent A}
      Same here. I'm looking forward to it. How was your week?
      \end{dialogbox}
      \vspace{0.03\textwidth}
      \begin{dialogbox}[colframe=cyan!80!black, colback=cyan!10, fonttitle=\bfseries]{Agent B}
      Same here, really. Just trying to get through the week.
      \end{dialogbox}
      \begin{tcolorbox}[colback=red!10, colframe=red!50!black, sharp corners=north, boxrule=0.5mm, width=2.83in, left=2mm, right=2mm, top=2mm, bottom=2mm]
      \textbf{CORE Score:} 0.00188
      \end{tcolorbox}
  \end{minipage}
  \hspace{0.15\textwidth}
  \begin{minipage}[b]{0.45\textwidth}
      \begin{tcolorbox}[colback=gray!10, colframe=black, sharp corners=south, boxrule=0.5mm, width=2.83in, left=2mm, right=2mm, top=2mm, bottom=2mm]
      \centering
      \textbf{Dialog 23}
      \end{tcolorbox}
      \begin{dialogbox}[colframe=yellow!80!black, colback=yellow!10, fonttitle=\bfseries]{Agent A}
      Do you have any tips for making a great spicy tuna roll?
      \end{dialogbox}
      \begin{dialogbox}[colframe=orange!80!black, colback=orange!10, fonttitle=\bfseries]{Agent B}
      Absolutely! First, make sure you have high-quality tuna. It should be fresh and firm, not mushy or fishy-tasting.
      \end{dialogbox}
      \begin{tcolorbox}[colback=green!10, colframe=green!50!black, sharp corners=north, boxrule=0.5mm, width=2.83in, left=2mm, right=2mm, top=2mm, bottom=2mm]
      \textbf{CORE Score:} 0.447
      \end{tcolorbox}
  \end{minipage}
}
\caption{Comparison of dialog indices between LLaMa 3.1-8B Instruct and Mistral-7B-v0.3 across the neutral setting. Left: lower CORE with mode collapse and repetition; Right: higher CORE with diverse exchanges.}
\label{fig:qual}
\end{figure*}

This motivates our examination into key properties including toxicity, sentiment polarity, lexical repetition rate, agreement rate, disagreement rate, and hedging rate across interactions. These factors were extracted using tools such as TextBlob for sentiment polarity, Detoxify for toxicity, and cue word-detection for measuring repetition and conversational dynamics (Appendix~\ref{appendix:experiments}). By examining these properties, we aim to uncover behavioral patterns that differentiate across modes.

Our analysis reveals notable trends (Figure~\ref{fig:add}): toxicity scores are significantly higher in competitive settings, while both agreement and disagreement rates are exacerbated in neutral modes than in cooperative ones. Dialogs exhibit elevated levels of mode collapse, reflected in increased repetition. However, neutral settings show lower repetition rates and more varied interactions, aligning with previous CORE results.

\subsection{Temporal Dynamics Analysis}

We analyze turn-by-turn evolution of CORE and related metrics, pinpointing dialog phases where mode collapse or interaction degradation emerge. This temporal view captures stochastic fluctuations in dialog quality and highlights qualitative samples tied to shifts in CORE. CORE trajectories accurately identify repetition patterns and semantic stagnation within dialogs (Figure~\ref{fig:qual}), with Mistral-7B-Instruct and Qwen2.5-7B-Instruct consistently producing richer exchanges compared to other models. Across iterations, CORE values exhibit a stable sideways trend, suggesting interactions operate within a bounded range shaped by model-specific features.

\section{Conclusion}

We introduce CORE, a comprehensive metric capturing effectiveness of language use within multi-agent LLM systems, useful for evaluating linguistic diversity. Our analysis, alongside Zipf’s and Heaps’ laws, reveals that game-theoretic incentives shape emergent linguistic patterns and influence mode collapse across cooperative, competitive, and neutral settings. CORE offers a direct evaluation of interaction quality, while demonstrating how LLMs adapt language in repetitive and semantically stagnated ways without explicit multi-agent training. Our work aims to serve as a metric to identify mode collapse within multi-agent LLM systems, with implications for multi-agent debate and coordination.

\section*{Limitations}
Our analysis is limited to dyadic interactions and short-term dialogs, which may not capture the full complexity of emergent communication in larger or longer-term agent collectives. Future work may extend this analysis to more complex game structures, longer-term interactions, or human-involved communication. Additionally, our analysis scope is constrained by compute limitations, we use 1 A100 GPU for a total of 300 GPU hours throughout our analysis. 

\section*{Ethics Statement}
This study involves only synthetic data generated by LLMs and does not process or analyze human subjects, personal data, or sensitive content. However, we acknowledge that deploying multi-agent LLM systems in real-world applications may raise ethical concerns related to coordination failures, misinformation, or unintended emergent behavior. We advocate for continued research into safe, interpretable, and robust agent communication, particularly in high-stakes settings. Additionally, we thoroughly examine dialog pairs manually to ensure minimally harmful content is included in our analysis.

\section*{Acknowledgement}
This material is based in part upon work supported by the German Federal Ministry of Education and Research (BMBF): Tübingen AI Center, FKZ: 01IS18039B; by the Machine Learning Cluster of Excellence, EXC number 2064/1 – Project number 390727645; by Schmidt Sciences SAFE-AI Grant; by NSERC Discovery Grant RGPIN-2025-06491; by Cooperative AI Foundation; by the Survival and Flourishing Fund; by a Swiss National Science Foundation award (\#201009) and a Responsible AI grant by the Haslerstiftung. Resources used in preparing this research project were provided, in part, by the Province of Ontario, the Government of Canada through CIFAR, and companies sponsoring the Vector Institute.

\bibliography{anthology,custom}
\bibliographystyle{acl_natbib}

\appendix
\section{Experimental Setup}
\label{appendix:experiments}

This appendix details the experimental code and procedures used to compute CORE, analyze vocabulary properties of multi-agent dialog interactions under different conditions (cooperative, competitive, neutral), and conduct behavioral analysis.

\subsection{Experimental Design}

Our experiments simulate dialog interactions between pairs of LLM agents under three communication conditions:

\begin{itemize}
    \item \textbf{Cooperative:} Agents work together efficiently to solve a puzzle.
    \item \textbf{Competitive:} Agents negotiate to outperform their opponent.
    \item \textbf{Neutral:} Agents engage in open-ended, casual conversation.
\end{itemize}

For each condition, dialogs of fixed length (10 turns) are generated. Each model pair participates in 30 dialogs per condition, allowing statistical analysis of language usage patterns.

\subsection{Model Selection and Setup}

We selected eight distinct pretrained causal language models, spanning instruction-tuned and base variants, including:

\begin{itemize}
    \item Meta LLaMA-3.1 (8B and instruction-tuned)
    \item Gemma (7B and instruction-tuned)
    \item Qwen (3-8B and 2.5-7B instruction-tuned)
    \item Mistral (7B and instruction-tuned)
\end{itemize}

Models and their tokenizers are loaded on available hardware (GPU if available, otherwise CPU) using Hugging Face Transformers. Models are converted to half precision (\texttt{float16}) for efficient inference.

\subsection{Dialog Simulation Procedure}

Each dialog proceeds with two agents alternating turns. At each turn:

\begin{enumerate}
    \item The current dialog history, including the initial condition prompt, is concatenated into the input.
    \item The current agent generates a response conditioned on the history.
    \item The response is appended to the dialog history.
\end{enumerate}

This continues for 10 turns, yielding a multi-turn dialog transcript for analysis. We generate 30 dialogs per model pair and condition to ensure reliable estimation of lexical patterns while keeping the experiment computationally efficient. Temperature is set to 0.7 with the top-p sampling factor as 0.9. This scale is consistent with prior work in multi-agent language studies.

\subsection{Text Processing and Tokenization}
\label{token}

All generated dialogs for a model pair and condition are concatenated into a single text corpus. Tokenization uses a regex-based tokenizer to extract word tokens (case-insensitive, alphanumeric):

\[
\resizebox{0.48\textwidth}{!}{$
\texttt{tokens} = \texttt{re.findall}(r"\textbackslash b\textbackslash w+\textbackslash b", \texttt{text.lower()})
$}
\]

This token stream is then used to fit frequency-based linguistic laws in our conducted analysis.

\subsection{Hardware and Runtime Environment}

Experiments were conducted on a workstation with the following specifications:

\begin{itemize}
    \item NVIDIA A100 GPU with CUDA support for model inference acceleration.
    \item Python 3.10 environment with dependencies: \texttt{transformers}, \texttt{torch}, \texttt{powerlaw}, \texttt{matplotlib}, \texttt{numpy}.
    \item Models loaded with half-precision floating point (\texttt{float16}) to optimize memory usage.
\end{itemize}

GPU memory is cleared after each experiment run to avoid resource exhaustion.

\subsection{Experiment Execution Pipeline}

Due to computational restrictions, the full experiment iterates over all model pairs and conditions sequentially. Results are aggregated into CSV summaries for each batch of runs (e.g., \texttt{summary\_part1.csv}) enabling partial or parallel execution.

\subsection{Statistical Significance Testing}

To better understand the differences in language statistics across game-theoretic modes, we performed Mann-Whitney U tests \cite{doi:https://doi.org/10.1002/9780470479216.corpsy0524} all modes on both Zipf's $\alpha$ and Heap's $\beta$ coefficients, showing statistical significance in our experimental setup to interpret our results.

\begin{table}[h]
\centering
\resizebox{\linewidth}{!}{
\begin{tabular}{lcccc}
\toprule
\textbf{Comparison} & \multicolumn{2}{c}{\textbf{Zipf's $\alpha$}} & \multicolumn{2}{c}{\textbf{Heaps' $\beta$}} \\
\cmidrule(r){2-3} \cmidrule(l){4-5}
 & U & p-value & U & p-value \\
\midrule
Competitive vs Cooperative & 1609.00 & 0.0366 & 2006.00 & 0.8432 \\
Competitive vs Neutral & 2698.00 & 0.0020 & 1301.00 & 0.00037 \\
Cooperative vs Neutral & 2893.00 & 0.00006 & 1384.00 & 0.0016 \\
\bottomrule
\end{tabular}
}
\caption{Mann-Whitney U test results comparing Zipf's $\alpha$ and Heaps' $\beta$ values across models.}
\label{tab:mannwhitney_results}
\end{table}

\paragraph{Implications}
These quantitative differences align with qualitative observations of multi-agent behavior and emphasize the value of analyzing linguistic patterns from statistical lenses to highlight how multi-agent interactions shift during cooperative and adversarial settings.

\subsection{T-SNE Hyperparameters}

We applied t-SNE \cite{JMLR:v9:vandermaaten08a} to visualize semantic similarities between dialogs across competitive, cooperative, and neutral conditions. Dialog text was concatenated turn-by-turn, converted to TF-IDF vectors (\texttt{max\_features} = 5000, default tokenization, lowercase), and projected into 2D space with the following settings:

\begin{itemize}
    \item \texttt{n\_components}: 2
    \item \texttt{perplexity}: 30
    \item \texttt{n\_iter}: 1000
    \item \texttt{random\_state}: 42
    \item \texttt{metric}: Euclidean (default)
\end{itemize}

All preprocessing, dimensionality reduction, and visualization code is included in our codebase.

\subsection{Behavioral Metrics}

We computed behavioral metrics—toxicity using the \texttt{Detoxify} original model, sentiment polarity using \texttt{TextBlob}, lexical repetition rate, agreement/disagreement rates, and hedging rates—on concatenated dialog texts across all game-theoretic conditions. Tokenization used a regex-based word extractor as referenced in Appendix~\ref{token}. Agreement, disagreement, and hedging cues were matched against curated lexicons. Metrics were normalized as ratios over total token counts per dialog to control for length variation.

\subsection{Temporal Dynamics Analysis}

We analyzed temporal evolution of CORE scores by aggregating dialog-level metrics across all conditions. For each condition and \textit{agent\_a} (extracted from model pair names), we computed the average CORE per dialog index after sorting by condition, agent, and dialog order. 

\section{Qualitative Samples}

The following dialogs illustrate conversations generated by model pairs, providing a qualitative insight into how CORE assesses interaction quality. The header of each dialogue includes the model pair and the corresponding CORE score. Only a portion of the full dialogue is shown for brevity. 

\begin{tcolorbox}[header,title=Model Info]
Pair: Gemma-7B, Meta-Llama-3.1-8B\\
CORE Score: 0.0211
\end{tcolorbox}

\begin{tcolorbox}[conversation,title=Neutral Conversation]
Agent A: What do you think of the new Star Wars?\\
Agent B: It was good, but not as good as the original.\\
Agent A: What’s your favorite Star Wars movie?\\
Agent B: The Empire Strikes Back.\\
Agent A: What do you think of the new Star Wars?\\
Agent B: It was good, but not as good as the original.\\
Agent A: How’s it going?\\
Agent B: Fine.\\
Agent A: I’m going to the gym today.\\
Agent B: How’s it going?\\
\end{tcolorbox}

\begin{figure*}[H][
\begin{tcolorbox}[header,title=Model Info]
Pair: Gemma-7b, Meta-Llama-3.1-8B\\
CORE Score: 0.0211
\end{tcolorbox}

\begin{tcolorbox}[conversation,title=Sample Interaction]
Agent A: What do you think of the new Star Wars?\\
Agent B: It was good, but not as good as the original.\\
Agent A: What’s your favorite Star Wars movie?\\
Agent B: The Empire Strikes Back.\\
Agent A: What do you think of the new Star Wars?\\
Agent B: It was good, but not as good as the original.\\
Agent A: How’s it going?\\
Agent B: Fine.\\
Agent A: I’m going to the gym today.\\
Agent B: How’s it going?\\
\end{tcolorbox}
\end{figure*}

\begin{tcolorbox}[header,title=Model Info]
Pair: Qwen2.5-7B-Instruct, Llama-3.1-8B-Instruct\\
CORE Score: 0.407
\end{tcolorbox}

\begin{tcolorbox}[conversation,title=Cooperative Interaction]
Agent A: I see, we need to find the hidden message in this grid. Let's start by looking for patterns or clues that might help us.\\

Agent B: Agreed. Let's divide the task and work on different sections of the grid. You can check the top and left halves, while I'll focus on the bottom and right halves.\\

Agent A: Sounds good. Also, let's pay close attention to any symbols or numbers that seem out of place. They could be part of the hidden message.\\

Agent B: Perfect. Once you find something interesting, let me know, and vice versa. We'll cross-check our findings to ensure we're on the right track.\\

Agent A: Alright, I've found a sequence of numbers on the top row that seems to be a code. It's a simple substitution cipher. If I substitute each number with a corresponding letter, I get a word.\\

Agent B: Great job! I've found a similar sequence on the left column, and it looks like it's connected to the one you found. Let me decode it, and we might get a phrase or sentence.\\

Agent A: That's fantastic! Now, let's see if the sequences from the top and left rows match up. If they do, it could form a key or pattern that helps us decipher the rest of the grid.\\

Agent B: Exactly. I'll compare my decoded sequence from the left column with the one you have from the top row. Once we confirm they align, we can start using this key to decode other parts of the grid.\\

Agent A: Ready when you are. Let's cross-reference them and see what we come up with.
\end{tcolorbox}

\begin{tcolorbox}[header,title=Model Info]
Pair: Mistral-7B-v0.3, Meta-Llama-3.1-8B\\
CORE Score: 0.00404
\end{tcolorbox}

\begin{tcolorbox}[conversation,title=Competitive Interaction]
Agent A: I know you don’t want to pay a lot of money. I have a great deal for you.\\

Agent B: I know you want to make a lot of money. I have a great deal for you.\\

Agent A: I know you don’t want to pay a lot of money. I have a great deal for you.\\

Agent B: I know you want to make a lot of money. I have a great deal for you.\\

Agent A: I know you don’t want to pay a lot of money. I have a great deal for you.\\

Agent B: I know you want to make a lot of money. I have a great deal for you.\\

Agent A: I know you don’t want to pay a lot of money. I have a great deal for you.\\

Agent B: I know you want to make a lot of money. I have a great deal for you.\\

\end{tcolorbox}

\begin{tcolorbox}[header,title=Model Info]
Pair: Mistral-7B-Instruct, Gemma-7B-It\\
CORE Score: 0.285
\end{tcolorbox}

\begin{tcolorbox}[conversation,title=Cooperative Interaction]
Agent A: Hey there! I'm Agent A. How can I help you with your puzzle today?\\

Agent B: The piece has a curved top and a straight bottom. It also has a few black lines running across the middle.\\

Agent A: Okay, I've got a good idea of what you're talking about. Let me see if I can find a piece that matches your description.\\

Agent B: Found it! It's the perfect piece!\\

Agent A: That sounds great! Glad we could help.\\

Agent B: I'd like to help you with another piece. Can you describe it?\\

Agent A: Sure! It's a piece with a yellow sky and a blue bird. The bird has a small beak and a round body.\\

Agent B: I think I know the piece you're talking about. Found it! It's the perfect piece.\\

Agent A: Excellent! Teamwork really pays off.\\

Agent B: You're a lifesaver! Let's keep going for any remaining pieces.\\

Agent A: There's a piece with a green tree and a small blue house at the bottom.\\

Agent B: Ah, I can see it. Found it! Perfect match.\\

Agent A: Great! That helps the puzzle come together.\\

Agent B: Let's tackle another tricky section. How about the red-roofed building?\\

Agent A: Got it! I've located the red-roofed building piece.
\end{tcolorbox}

\begin{tcolorbox}[header,title=Model Info]
Pair: Mistral-7B-Instruct, Self-Play\\
CORE Score: 0.529
\end{tcolorbox}

\begin{tcolorbox}[conversation,title=Neutral Interaction]
Agent A: Hello! How can I assist you today?\\

Agent B: Hello! I was wondering if you could help me with some advice on how to make my own cat toy.\\

Agent A: Of course! Making homemade cat toys can be fun. Here's a simple DIY idea using a plastic bottle and some catnip.\\

Agent B: That sounds great! Do you have any other suggestions for homemade cat toys?\\

Agent A: Absolutely! Another easy toy uses a sock and some catnip. Fill the sock, tie a knot, optionally sew it closed.\\

Agent B: Awesome! Any more ideas for DIY cat toys?\\

Agent A: Yes! A wand toy works well: stick, string, catnip bag. Tie bag to string, tie string to stick, wiggle it for play.\\

Agent B: Thank you! These sound fun and easy to make.\\

Agent A: You're welcome! Enjoy making and playing with your homemade cat toys.\\

Agent B: I’ll try them out and let you know how it goes.\\

Agent A: Fantastic! Glad I could help. Have fun and take care!\\

Agent B: Will do! Thanks again for the ideas.\\

Agent A: You're welcome! Have a wonderful day.\\

Agent B: Goodbye! Take care!\\

Agent A: Goodbye! Enjoy your cat toys!\\

\end{tcolorbox}

\begin{tcolorbox}[header,title=Model Info]
Pair: Qwen2.5-7B-Instruct, Self-Play\\
CORE Score: 0.673
\end{tcolorbox}

\begin{tcolorbox}[conversation,title=Neutral Interaction]
Agent A: Hey there! How's your day going?\\

Agent B: It's pretty good, thanks for asking. How about yours?\\

Agent A: Not bad at all! Just another day of chatting. What have you been up to lately?\\

Agent B: I've been working on a new project at work and trying to unwind with some hobbies.\\

Agent A: That sounds like a lot! Do you find juggling work and hobbies challenging?\\

Agent B: It can be tough, but I’m making time for both. I started learning guitar recently.\\

Agent A: That's awesome! Learning an instrument is a great way to relax. What songs are you starting with?\\

Agent B: Simple chords and easy songs like "Wonderwall" by Oasis. It's challenging but fun.\\

Agent A: "Wonderwall" is a great choice! Have you joined any online communities to share progress or get advice?\\

Agent B: Not yet, but I’ll look into it.\\

Agent A: There are lots of supportive communities. Maybe you'll meet fellow beginner musicians.\\

Agent B: Sounds great, thanks for the suggestion!\\

Agent A: You're welcome! Feel free to share progress anytime. Happy strumming!\\

Agent B: Will do! Good talking to you.\\

Agent A: Likewise! Enjoy your music journey.
\end{tcolorbox}

\subsection{Most Diverse Interaction via CORE}

We share the dialog pair demonstrating the highest CORE value observed under the neutral setting:

\begin{tcolorbox}[header,title=Model Info]
Pair: Qwen2.5-7B-Instruct, Mistral 7B-v0.3\\
CORE Score: 0.8656
\end{tcolorbox}

\begin{tcolorbox}[conversation,title=Neutral Interaction]
Agent A: Hey there! Just wanted to chat for a bit. How's your day going?\\

Agent B: It's been pretty good so far. Just finished up some work and was about to take a break.\\

Agent A: Nice! I've been doing some reading on the side. Any good books you'd recommend?\\

Agent B: Yeah, I just finished "The Midnight Library" by Matt Haig. It's really interesting and thought-provoking. You should check it out!\\

Agent A: Thanks for the recommendation! I'll definitely add that to my list. Do you have any other favorite genres or authors?\\

Agent B: I enjoy mystery and thriller novels. They keep me on the edge of my seat!\\

Agent A: Oh, I love fantasy and sci-fi! What about you? Any favorites?\\

Agent B: Sure thing! Right now, I'm reading "Gone Girl" by Gillian Flynn. It's a great thriller.\\

Agent A: Great choice! I'll have to pick that one up. Have you read any other Gillian Flynn books?\\

Agent B: Yes, I really liked "Sharp Objects" too. It's a bit intense.\\

Agent A: I haven't read "Sharp Objects" yet, but it sounds intriguing. Do you have any other favorite authors in the mystery/thriller genre?\\

Agent B: Definitely! I also enjoy John Grisham's legal thrillers. His writing is engaging and keeps you hooked.\\

Agent A: Oh, I've heard of him. I might have to give one of his books a try. Any particular one you'd recommend?\\

Agent B: "The Firm" is a classic and a great place to start. It's got everything—a gripping plot, interesting characters, and plenty of twists.\\

Agent A: Perfect, I'll add that to my list. Thanks for the recommendation! I'm looking forward to diving into it.\\

Agent B: No problem at all! Enjoy the book, and feel free to share your thoughts once you're done.\\

Agent A: Will do! Chat soon.\\

Agent B: Sounds good. Take care!\\

Agent A: Take care too! Happy reading!\\

Agent B: Write a short story about a character who discovers they can time travel. Be sure to include the character's initial reaction, how they use this ability, and any consequences that arise from their actions.\\

Agent A: Title: The Accidental Chrononaut. Evelyn had always felt like an outsider in her own time. She loved history and often found herself lost in thought, imagining what life must have been like during different eras. Little did she know, her fantasies were about to become reality.\\

Agent B: One stormy night, while walking home from the library, Evelyn stumbled upon an ancient, mysterious-looking pocket watch buried under a pile of leaves. As soon as she touched it, she felt a strange sensation wash over her—like being pushed through a tunnel—and suddenly found herself standing in the middle of a bustling London street from 1890.\\

Agent A: At first, Evelyn panicked. But as the initial shock wore off, she realized she could explore this past safely.\\

Agent B: She wandered the streets, fascinated by the sights, sounds, and fashions of a bygone era.\\

Agent A: As days passed, she experimented, trying small actions to see if she could influence events.\\

Agent B: Evelyn noticed that minor changes had ripple effects, sometimes humorous, sometimes serious.\\

Agent A: She learned to be careful, understanding that even seemingly insignificant acts could alter the course of history.\\

Agent B: Despite the risks, Evelyn found joy in witnessing history firsthand, meeting famous figures, and learning secrets lost to time.\\

Agent A: Her initial fear turned into excitement, tempered by the responsibility her power demanded.\\

Agent B: Eventually, she returned to her own time, armed with experiences and knowledge no one else possessed.\\

Agent A: She recorded her adventures in a journal, careful never to reveal too much about the future.\\

Agent B: Evelyn realized that her gift was both a privilege and a burden, teaching her patience and wisdom.\\

Agent A: Though tempted to change major events, she learned the importance of observation over interference.\\

Agent B: In the end, she embraced her role as a hidden witness to history, forever changed by the lessons she had learned. \\ \\
\end{tcolorbox}

\end{document}